\pdfoutput=1

\documentclass[discussion]{clv3}

\usepackage[T1]{fontenc}

\usepackage[utf8]{inputenc}

\title{Do Language Models' Words Refer?}
\runningtitle{Do Language Models' Words Refer?}
\runningauthor{Matthew Mandelkern and Tal Linzen}

\author{Matthew Mandelkern}
\affil{Department of Philosophy\\New York University\\ \texttt{mandelkern@nyu.edu}}

\author{Tal Linzen}
\affil{Department of Linguistics and
  Center for Data Science\\
  New York University\\
  \texttt{linzen@nyu.edu}}

\begin{document}
\maketitle
\begin{abstract}
What do language models (\emph{LM}s) do with language? Everyone agrees that they can produce sequences of (mostly) coherent strings of English. But do those sentences \emph{mean} something, or are LMs simply babbling in a convincing simulacrum of language use?  Here we will address one aspect of this broad question: whether LMs' words can \emph{refer}, that is, achieve ``word-to-world'' connections. There is prima facie reason to think they do not since LMs do not \emph{interact} with the world in the way that ordinary language users do. Drawing on insights from the externalist tradition in philosophy of language, we argue that those appearances are misleading: even if the inputs to an LM are simply strings of text, they are strings of text \emph{with natural histories}, and that may suffice to put LMs' words into referential contact with the external world.

\end{abstract}

\section{Introduction}

Language models (\emph{LM}s) are machine learning systems trained to capture the statistical distribution of sequences of words in a collection of texts. Once trained, these systems can be used to generate new texts drawn from the same distribution. 
When generating those texts, these systems, especially those based on neural networks trained on large amounts of text, often \emph{appear} to be producing series of meaningful sentences.
But some have argued that LMs are really nothing more than babbling ``stochastic parrots'' (a term due to \citealt{stochasticparrots}). One might think, in particular, that LMs produce a mere simulacrum of meaningful human language, because their words never achieve the referential connections required for meaning \citep{Ostertag:2024,Titus2024-TITDCH}.

This skepticism can be motivated in a number of ways. Here we will address one  argument, which says that because LMs' only input is strings of symbols, they cannot produce referential words: \emph{reference} cannot be derived from \emph{form}.
Here we will show that this argument is mistaken: LMs' input is not bare strings of symbols, but strings of symbols with certain natural histories which connect them to their referents. We  draw on the externalist tradition in the philosophy of language, which has convincingly argued that reference is  achieved, not by having particular beliefs about, experiences with, or capacities involving word referents (the kinds of things that LMs most plausibly lack) but rather by standing in the right kind of natural history to the referents. Hence the natural histories of LMs' inputs may be enough to ground reference for the LMs' outputs, even if the LMs have no way to grasp or access those histories.

\section{Our question}

To help get clear on the question we are asking, consider a contrast based on \citealt{Putnam:1981}. Suppose that, at a picnic, you observe ants wending  through the sand in a surprising pattern, which closely resembles the English sentence  `Peano proved that arithmetic is incomplete'. 
At the same time,  you get a text message from  Luke, who is taking a logic class. He writes, `Peano proved that arithmetic is incomplete'. 

Intuitively, the two cases are very different, despite involving physically similar patterns. The ants' patterns do  not \emph{say} anything; they just happen to have formed patterns which \emph{resemble} meaningful words. 
Of course, \emph{you} can interpret the pattern, just as you can interpret an eagle's flight as an auspicious augur; but these are interpretations you overlay on a natural pattern, not meanings intrinsic to the patterns themselves.
By contrast, Luke's words \emph{mean something definite} on their own (regardless of whether you or anyone else interprets them): namely, that Peano proved that arithmetic is incomplete. What Luke said is false: it was G\"odel who proved  incompleteness. But Luke \emph{said} something, whereas the ants didn't say anything at all. In particular, he said something false \emph{about Peano}, which means that his (use of the) word `Peano' managed to \emph{refer} to Peano.

We are interested in the question of whether the outputs of LMs' are more like the ants' patterns or  like Luke's text: do they merely \emph{resemble} meaningful sentences, or are they in fact meaningful sentences? Hence suppose that an LM produces the sentence `Peano proved the incompleteness of arithmetic'. Does this amount to the production of a \emph{false but meaningful} sentence, like Luke's, or is it a meaningless pattern which merely resembles a meaningful sentence, like the ants'?

There are many ways we could make this question precise; here, we will address the question of whether the words that LMs produce \emph{refer}, that is, achieve the word-to-world connections that ground meaningful language use. For instance, we can ask whether the word `Peano' output by the LM refers to Peano, or rather doesn't refer at all; we'll use \emph{referentiality} for the thesis that LMs' words do refer.\footnote{We  use `words' to describe these outputs, even though on one legitimate use of `word', a word is a \emph{meaningful} string or phoneme of a language. We  use `word' in the thinner but more familiar sense of an instance of a (potentially meaningless) string of symbols, so that it is uncontroversial that LMs output words, and controversial only whether those words are meaningful. One could reformulate our question in terms of whether LMs truly output words in the first sense \citep{Kaplan1990-KAPW}.} This is intimately related to the question of whether the sentence the LM produces is meaningful, since meaning (nearly all agree) requires reference.

There are many important questions to ask about LMs, language, and meaning.\footnote{For a helpful guide to philosophical questions about reference, see \citet{HawthorneManley}.}
Our question is just one. A closely related question to ours, which we will not tackle head-on, is whether the LM \emph{itself} refers (whether it achieves \emph{speaker reference}), which can in principle come apart from the question of whether \textit{its words} refer \citep{KripkeSpeakers}. 
While our question is  narrow, it strikes us as being of obvious importance to know whether LMs are more like ants in the sand, producing strings that happen to resemble human language, or are rather producing meaningful strings of human language. 

Our goal in this short paper is to bring arguments to the attention of the AI and NLP communities  from the externalist tradition in the philosophy of language  \citep{Kripke:1980,PutnamMM}. These arguments  undermine a very tempting argument that LMs' words cannot refer because LMs' inputs are only strings of symbols and they do not interact directly with the external world. However, we make no claim to a decisive argument that LMs' words \emph{do} refer.
There are many subtle issues in the neighborhood of those we address, and much more to say about every issue we raise. Our goal is to briefly explain externalism and begin to explore its ramifications for LMs in an accessible way that we hope will spur further exchanges and explorations of these ideas between the philosophy and AI/NLP communities.

To our knowledge, our paper is the first to  introduce the  ideas motivating externalism in a brief and accessible way to an AI/NLP audience. However, it is not the first to connect externalism to the question of referentiality.
\citet{Cappelen:2021} defend referentiality on externalist grounds, but argue that
standard externalist accounts don't apply to LMs because of `crippling anthropocentric biases'. By contrast, we will argue that standard externalist accounts are at least \textit{prima facie} compelling with respect to LMs.
\citet{Butlin:2021} argues that  externalism vindicates referentiality for chatbots but \emph{not} for LMs. 
\citet{Devault:2006} argued that externalism did not vindicate referentiality for NLP as it existed at the time, but could in principle apply. 
The accounts most similar to ours in viewing standard externalism as providing a possible vindication of referentiality are the excellent recent papers \citet{Pavlick:2023} and \citet{Mollo:2023}. We are very sympathetic to both discussions of externalism, but neither one explains the original motivations behind externalism and how those same motivations might apply to LMs, a lacuna we aim to rectify here.

\section{The grounding problem\label{grounding}}

LMs \emph{appear} to produce meaningful strings. Suppose an LM  outputs the string `Apples are fruits'; that string appears to be saying that apples are fruits, and hence using `apples' to refer to apples.
Still, you might think that the LM's output is no more intrinsically meaningful than figures accidentally sketched by ants in the sand because the LMs' use of language is not \emph{grounded} in beliefs, experiences, and capacities involving the external world. After all, an LM trained on language alone has had no sensory contact with apples; it has never tasted or touched or smelled them. It cannot sort apples from oranges. You might also think that LMs lack any desires and beliefs about apples.
Without a doubt, LMs do not have the store of apple-related memories and experiences that ordinary speakers of English generally have. So in the absence of relevant experiences, beliefs, desires, and capacities, how could they refer to apples?  LMs are just too isolated from the world for their symbols to be connected to it in the way required for reference.

Another way to put the point is  that \emph{form} does not suffice to ground \emph{reference}. Everyone will agree that the way that symbols are arranged together does not determine the referent of the arrangement. But, the thought goes, \emph{form is all that LMs have}, and so LMs cannot achieve reference.\footnote{ \citet{Bender:2020} make an argument with a similar form, but it differs in subtle ways: first, they are concerned with relating forms to a broader notion of meaning than just reference; and second, their main argument is that {training} on form is insufficient for an LM to \emph{learn} such relations, whereas our focus is simply on whether LMs' words \emph{bear} the reference relation.}

\section{Externalism about reference}

Drawing on the externalist tradition in philosophy of language, we will argue that this objection to referentiality is misplaced. What grounds  reference is not  beliefs, experiences, and discriminatory capacities of the kind that LMs obviously lack. Rather, what grounds reference is the \emph{natural histories} of words: the causal-historical links between a speech community's use of a word and the word's referent.\footnote{As a reviewer points out, we are using the term ``speech community'' in the idealized and somewhat underspecified sense it is used in the philosophical literature; for a critical discussion of this term from an empirical sociolinguistic perspective, see, e.g.,  \citet{Patrick:2004} and \citet{Morgan:2014}.}

Recall Luke. Suppose that he had never heard of Peano before recently reading a very short article about him, comprising just the sentence `In 1931, Peano proved that arithmetic is incomplete'. This is the first that Luke has heard of Peano, and hence amounts to his entire body of information about Peano.  

Now, as a matter of fact, this sentence is false. Nonetheless, just by being exposed to this sentence, Luke comes to be able to use the word `Peano' to refer to Peano. How do we know? Well, if he texts you `Peano proved that arithmetic is incomplete', his text \emph{says something false}, namely something \emph{about Peano}: that he proved incompleteness. So Luke's use of the word `Peano' refers to Peano.
This is  striking, since Luke's only substantive belief about Peano is the false belief that he proved that arithmetic is incomplete. Luke has nothing by way of experience with Peano, no fluency with regards to facts about his life, indeed, no ability to discriminate him from G\"odel. 

Of course, he has some trivial beliefs: that Peano is the referent of `Peano', that Peano is called `Peano' by Luke's speech community, and so on---something we'll return to below. There are other things you might think that he could infer from the sentence in question---for instance, that `Peano' was called  Peano by his contemporaries. But, as \citet{Kripke:1980} points out in an analogous discussion, we can easily suppose this to have been false: for instance, suppose that both Peano and G\"odel were called `Georg Kreisel' by their contemporaries,  and only later dubbed `Peano' and `G\"odel', respectively. Luke would still be able to refer to Peano with `Peano'. 

One lesson of thought experiments like this is that it is strikingly easy for a speaker to use a word referentially. This does not require substantive experience, beliefs, or capacities vis-\`a-vis its referent. This runs contrary to a prima facie compelling picture on which reference is achieved by having a certain body of information and experience with the referent. That picture was long popular in philosophy but was refuted by externalists like Kripke and Putnam; and we think that now-defunct idea underlies the argument about grounding sketched in the last section.

\subsection{Natural history and deference}

How does Luke's use of `Peano' come to refer to Peano, if not via Luke's beliefs about and experiences with Peano?  There is much to say about this question. We will briefly summarize two observations. 
The first is that a word use's \emph{natural history} can largely determine its reference. This is apparently what goes on with Luke's use of `Peano': the fact that Luke is part of a speech community whose history of using the word traces back to Peano, rather than G\"odel, seems to be enough to make the referent of the word in Luke's mouth Peano, rather than G\"odel---even though Luke's \emph{beliefs} about the referent of `Peano' uniquely pick out  G\"odel, not Peano.

For a striking illustration of the point, consider  \citeauthor{PutnamMM}'s famous \emph{Twin Earth} case. Compare speakers of English in an era when they do  not yet know the chemical composition of water to denizens of an alternate reality, Twin Earth, which is identical to earth in all macroscopic properties, but where the chemical composition of the stuff that in every macroscopic way resembles water is not H$_2$O but rather XYZ. By hypothesis, there are people on Twin Earth who speak a language which is superficially indistinguishable from English. But consider what their word `water' refers to: it refers to the stuff \emph{in their reality} which they call `water', which is XYZ, not H$_2$O. To see this, note that we Earthians can truly say, in our dialect, `There is no water on Twin Earth; instead, it contains a macroscopically identical substance, made of XYZ'. Conversely, Twin Earthians could, in \emph{their} version of English, truly say `On Earth there is no water, only H$_2$O'. 

What makes it the case that `water' in  Earthian English refers to H$_2$O, while `water' in Twin Earthian refers to XYZ?  It is \emph{not} the beliefs of the speakers about the liquids' respective chemical compositions, since, by assumption, they do not yet know the chemical compositions of their respective watery stuffs. (You could argue that the meanings of the word \emph{change} when the speakers learn the chemical compositions, but this is obviously wrong: scientists discovered \emph{that water was H$_2$O}, they did not change the meaning of `water'.)
Rather, it seems to simply be the fact that the stuff in the natural history of the word `water' on earth was made of H$_2$O, whereas the stuff in the natural history of the word `water' on Twin Earth was made of XYZ. 

There are different ways to spell out the notion of natural history here. \citet{Kripke:1980} talks evocatively of an initial baptism event, where a word is associated with a referent (`Let's use this word `water' to refer to this stuff over here'), which is connected to current usage by myriad causal-historical chains. That is obviously an idealization; the origin stories of word reference can be diffuse and complicated (see \citealt{VOR} for a different approach). But it provides a helpful schematic picture of how words can refer in ways that go beyond or even contradict the speaker's representation of the referent. 

In other cases, what  determines reference is \emph{deference} to contemporaneous experts. Consider another case of Putnam's: an ordinary speaker's use of the words `elm' and `beech'. Most speakers of English (let us suppose) don't know the difference between elms and beeches:  they don't have beliefs that distinguish the two types of trees; nor could they distinguish them if they were asked to; nor, in some cases, have they had any experiences of either. Still, `beech' in the mouth of such speakers refers to beeches, not elms; while `elm' in the mouth of an ordinary speaker refers to elms, not beeches. To see this, suppose such a speaker pointed to a tree which happened to be an elm, and said `This is a beech': we would take them to be saying something false, namely that the tree before them is a beech. So `beech' still refers to beeches in their mouth. Likewise for elms. 

How can this be? Part of the story is, again, the natural history of the word: `beech' in this speech community has historically been used to refer to beeches, and that is enough for ordinary speakers to use it that way, even if their beliefs and experiences don't connect them to beeches in any particular way. Another part of the story, Putnam argues, is \emph{deference}: ordinary speakers use `beech' to refer to \emph{whatever tree experts take it to refer to}, and that is enough to refer to beeches.

\subsection{The upshots for LMs}

Luke's use of `Peano' refers to Peano, not because of his beliefs about Peano, but  thanks to the fact that he read a line of text about Peano, and hence became part of a linguistic community whose use of `Peano' traces back via causal-historical lines to Peano himself. 
By contrast, the ants' pattern does not refer to Peano, because they aren't part of such a speech community. 

Now back to LMs. Suppose that an LM is trained on the very same text about Peano that Luke saw, and then generates the (false) sentence `Peano proved the incompleteness of arithmetic'.
Does the LM's word `Peano'  succeed in referring to Peano? 

If you thought that the difference between the ants and Luke was that Luke's words, but not the ants', are grounded in substantive beliefs, capacities, and experiences involving Peano, then you should think that LMs are more like ants (the skeptical position described in \S\ref{grounding}). But we have now seen  strong reason to think that this is \emph{not} what accounts for the difference; rather, the difference is that Luke, but not the ants, is part of a linguistic community whose use of `Peano' traces back to Peano. 

So is the LM part of a linguistic community whose uses of  `Peano' trace back to Peano? This is obviously a tricky question to answer, and we don't have decisive considerations one way or another. In the next section, we will consider some reasons to think the answer is `no', and argue that none of them is especially convincing.  But first, let us emphasize that \emph{this} is the central question to ask, for if the answer is yes, then LMs' use of `Peano' refers to Peano: no beliefs, capacities, or experiences with Peano are required for reference in general, and so there is no particular worry in the case of LMs either. 

One way to formulate the skeptical argument from grounding is that LMs have access only to form, and form underdetermines reference. Now we can see what is wrong with this argument. \emph{The inputs to LMs are not just forms, but forms with particular histories of referential use.}  And those histories ground the referents of those forms, whether or not you know them or have any kind of access to them.  Luke wrongly believes that the causal history of his use of `Peano' traces back to the discoverer of the incompleteness proof; nonetheless, the simple fact that the \emph{actual} history of `Peano' traces back to Peano, not G\"odel, suffices to make Peano, not G\"odel, the referent of Luke's use of `Peano'.

\section{Reasons for skepticism}

The key question for referentiality is thus not about grounding but rather whether LMs are part of a linguistic community whose use of `Peano' traces back to Peano. Unlike in the case of grounding---where a strong case could be made that LMs lack at least the relevant experiences and capacities, and perhaps also the beliefs---prima facie considerations tell in favor of counting contemporary neural network LMs such as ChatGPT as part of our linguistic community: namely, they speak and interact in linguistically ordinary ways. 
On the spectrum between ants and Luke, they \emph{look} much closer to Luke. 

This is, of course, only prima facie evidence: there might be compelling reasons to conclude that LMs are \emph{not} part of a speech community in the relevant sense.\footnote{From the other direction, as a reviewer points out, you might think that LMs are part of \emph{many} speech communities, corresponding to the diversity of their training data. That seems right to us, though a similar point applies, at a smaller scale, to humans. This, in turn, might introduce a great deal of \emph{indeterminacy} in LMs' reference. But, again, the same plausibly applies (on a smaller scale) to human reference.} We won't try to argue decisively against this position. But we will address some of the most obvious arguments and argue that they are less convincing than they might first appear.

One tempting thought is that being part of a speech community that uses a word to refer in a certain way---and thus being able to use that word referentially---requires being disposed to draw the particular inferences from that word that are drawn by members of that community. So, for instance, even if you can't distinguish elms from beeches, there are lots of things that you can still infer from `This is an elm', for instance, `This is a tree', or `This is an object'. Similarly, even though Luke doesn't know anything about Peano, he is likely to infer certain things about him, based on his general knowledge about logic and the world. Clearly inferences aren't \emph{enough} to determine reference. Earthians and Twin Earthians will associate the same inferences with `water', but the words in their respective languages refer to different things. But inference drawing of some kind may still be a necessary condition on using a word to refer (see e.g. \citeauthor{PutnamMM}'s \citeyear{PutnamMM} discussion of stereotypes). 

But it is hard to see how the capacity to draw appropriate inferences would be a problem for the claim that LMs can refer. What seems potentially hard for LMs is ``word-to-world'' connections, since LMs are in some obvious sense isolated from the external world. By contrast, ``word-to-word'' connections, and in particular inferences, should in principle be possible for a language model to acquire, since connections between words just are its stock in trade (see \citealt{Potts:2020,Merrill:2022,piantadosi2022meaning} for relevant discussion). 

Another reason for skepticism you might have is that LMs don't have the right kinds of \emph{intentions} to produce referring words. The thought is that to produce an instance of `Peano' that refers to Peano, the speaker must intend to use `Peano'  to refer to Peano, and that is something LMs can't do.

There is a thick and a thin version of this thesis. The thick version is that a speaker can produce a word which refers to x iff she intends to use it to refer to whatever verifies some cluster of her substantive beliefs about x. This is a tempting picture, but we have seen that it is refuted by the externalist considerations we've seen so far. Luke intends to use `Peano' to refer to whoever proved that arithmetic is incomplete (since he intends to speak truly); but he in fact succeeds in using `Peano' to refer to \emph{Peano}, not G\"odel. Speaker intentions in this thick sense are thus neither necessary nor sufficient for reference. 

A thinner version of the view is more plausible, given externalist considerations: a speaker can produce a referential word  iff she intends that word to refer to whatever its natural history in her speech community determines as its referent (cf. \citealt{GriceMeaning,Clark:1996} for ideas along these lines).
Again we can consider more and less demanding versions of this thinner view. On a more demanding version, referential production requires high-level reflection about the notion of linguistic community and reference; you need to think something like, `I'm using this word to refer to whoever `Peano' refers to in my speech community' in order to produce a referential instance of `Peano'. That is obviously wrong; obviously, many referentially successful speakers never think \emph{explicitly} about reference or speech communities. 
A less demanding version says that what is required is only some kind of \emph{implicit}
representation of oneself as part of a speech community and intention to refer in conformity with the word's natural history in that speech community---the kind of implicit representation that can be ascribed to, say, a child who successfully produces referential words, but may be unable to speak coherently about reference or speech communities.

Then the key question is whether LMs can have this kind of lightweight intention to refer in conformity with a language community. To be clear, although lightweight, this is not a trivial criterion to meet. Ants obviously don't meet it; Luke obviously does. LMs are  a  much harder case, and it would take a lot of work in philosophy of mind to explore whether they can intend to refer in conformity with their speech community in this implicit, sub-personal way. 
Certainly ordinary speakers are happy to talk about machines acting intentionally (we'll happily say that a robotic arm intends to pick up a cup of water). And it is plausible to argue that LMs could, and increasingly do, learn to model communicative intention in order to accurately predict upcoming words \cite{andreas-2022-language}. But there is still plenty of room to deny that LMs intend to refer in line with their linguistic community (see \citealt{Ostertag:2024} for one development of this position). For instance, you might think that intending requires consciousness, or theory of mind, or self-representation, or intrinsic goals, or something else, and then argue that LMs lack these things.\footnote{See \citet{Chalmers:2024} for recent discussion of whether LMs, present or future, are conscious.}
We leave this as an open question. But in each case, we warn against holding LMs to a higher standard than we hold ordinary speakers, like children, who may not have clear representations of their speech community, and yet we are happy to ascribe referential success to those speakers.

\section{Conclusion}
Perhaps LMs are like ants formicating meaninglessly in the sand, producing mere simulacra of  referential language. 
This is a natural conclusion if you think that reference is achieved by connecting a word to whatever it is that satisfies a body of substantive beliefs, experiences, and capacities involving the referent. Externalist arguments in the philosophy of language, however, have shown that reference does not work that way: reference turns out to be surprisingly disconnected from speakers' beliefs, and surprisingly easy. Simply being part of a speech community whose use of a word is historically connected to a particular referent in the right way suffices for a speaker to be able to produce a referring instance of that word. 

This helps vitiate a natural worry about referentiality, the worry from grounding. Even if LMs' inputs are only strings, those strings have a natural history which connects them  to referents; and those natural histories are, in principle, enough to make their words refer. 
Of course, there are many things that LMs can't do. They may be totally incompetent with respect to the substantive facts about a given domain. But that doesn't obviously stop them from referring, any more than Luke's erroneous beliefs about Peano stops \emph{his} use of `Peano' from referring to Peano. 

As we have seen, there might still be reasons to doubt that LMs refer based on doubts about whether they can really be seen to be part of a speech community; but it is \emph{this} question that is crucial for the question of referentiality, not questions about  whether LMs have the right kinds of experiences, beliefs, desires, and so on to ground reference.

Even if you conclude that LMs are not part of our speech community, this does not foreclose the possibility that their words still refer. For instance, \citet{Lederman:2024} argue that LMs are not part of our speech communities, but their words nevertheless refer because they are causally sensitive to the  \emph{intelligibility} of their data. For all we've said, this position, or yet another one, could turn out to justify the claim that LMs' words refer, even if a more straightforward externalist vindication of referentiality along the lines we have explored here ultimately fails.

Let us close with a thought experiment. Suppose that Izzy is isolated at birth in a sensory deprivation chamber; his only contact with the outside world is via an interactive screen, directly sensitive to his brainwaves, which displays text and can send back messages. Amazingly, Izzy gradually comes to be an apparently competent user of the screen. He can send messages of varying degrees of sophistication and correctness, like `Get me out of here!' or `Peano proved that arithmetic is incomplete', and he can interact with inputs on the screen in linguistically fairly normal ways. Presumably most people's first reaction would be that Izzy has, amazingly, become a part of our speech community, and that when he writes out something like `Peano proved that arithmetic is incomplete', he has managed to refer to Peano. 
On the face of it, there is a deep puzzle here about how his words could refer to the external world, given Izzy's isolation from it. But externalism teaches us that there is not really a puzzle here: Izzy's words can refer to the external world even though he is profoundly isolated from it, because reference is grounded in the causal-historical chains running through a speech community connecting a use of a word to its referent. In the same way, externalism dissolves the same puzzle about how an LM's use of language could be referential, despite its isolation from the external world.

\section*{Acknowledgments}

We thank Justin Bledin, David Chalmers, Josh Dever, Thomas Icard, Cameron Kirk-Giannini, Will Merrill, Alexander Koller, Harvey Lederman, Christopher Potts, Samuel Rogers, Daniel Rothschild,  David Schlangen, Sebastian Schuster, and three anonymous referees for this journal for very helpful discussion and feedback on earlier drafts.

\bibliography{bibliography}
\bibliographystyle{compling}

\end{document}